\documentclass[]{dice}
\definecolor{metablue}{HTML}{1772B4}

\setabstractbrandtext{XMed Lab}
\setabstractbrandcolor{metablue}
\setabstractbrandlogo{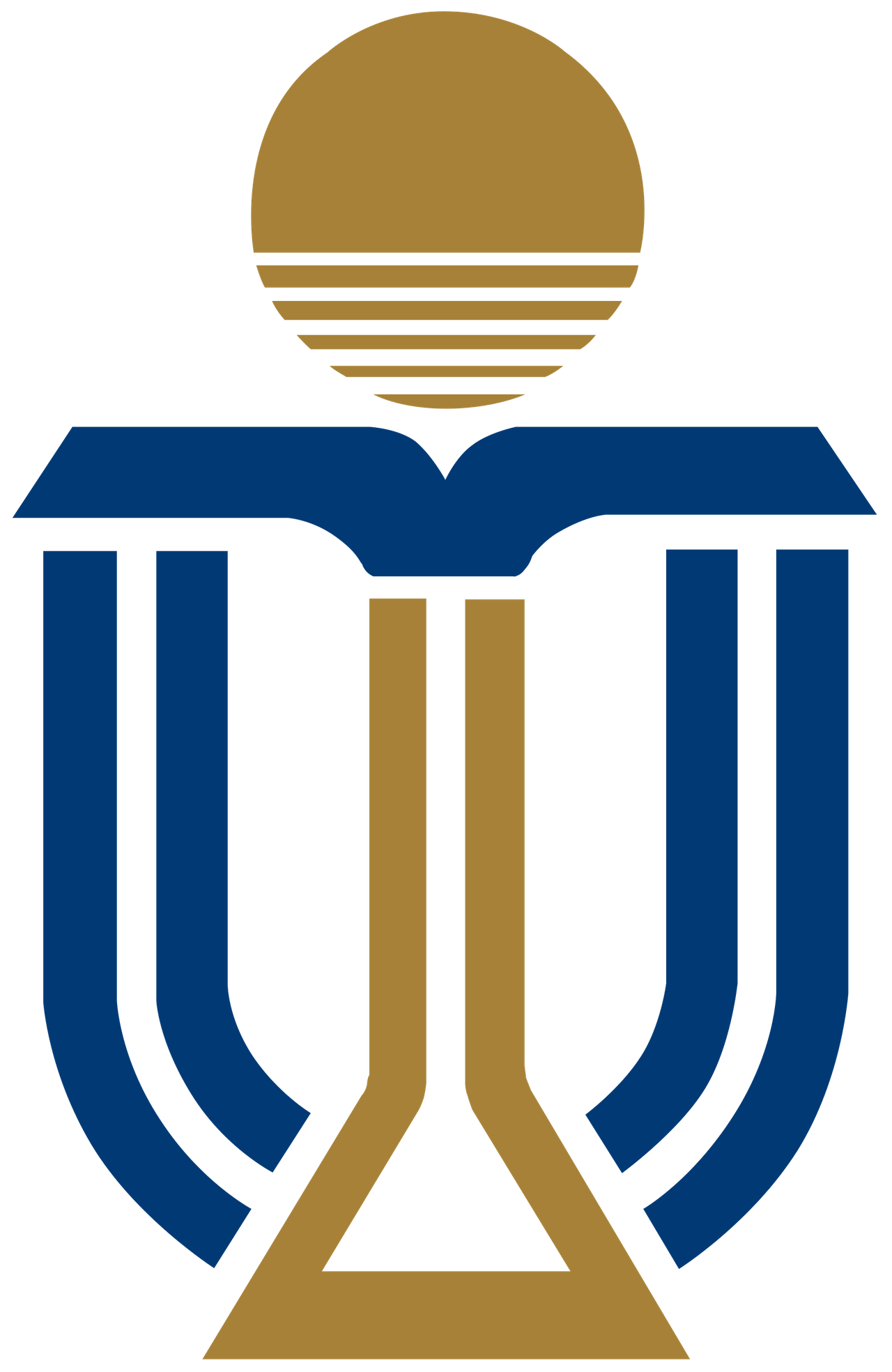}
\setabstractbrandlogosecond{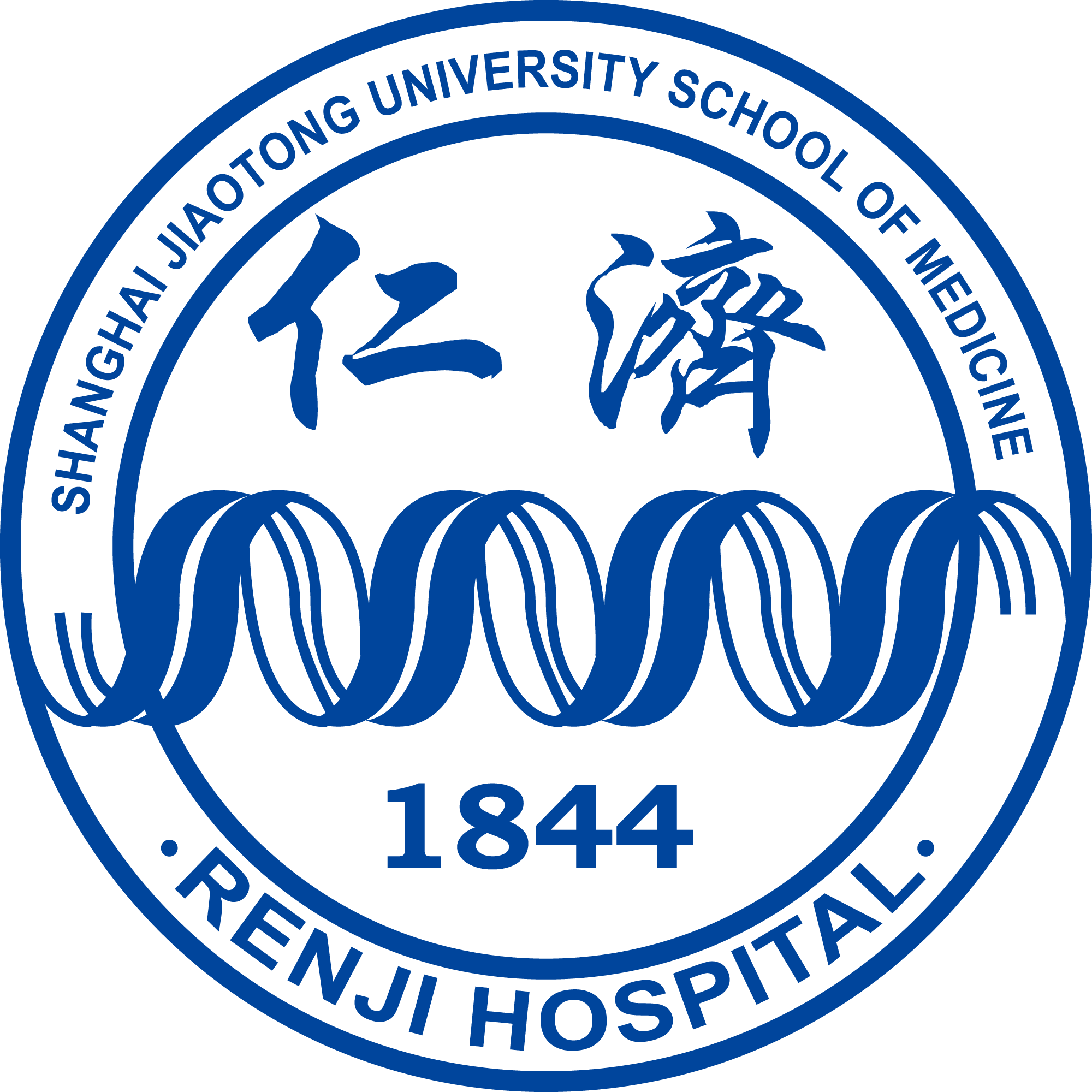}
\setabstractbrandlogoheight{1.2cm}

\usepackage{amsmath,amssymb,amsfonts}
\usepackage{mathtools}
\usepackage{amsthm}
\usepackage{bbm}
\usepackage{bm}

\usepackage{subfigure}
\usepackage{float}
\usepackage{adjustbox}
\usepackage{colortbl}
\usepackage{arydshln}

\usepackage{algorithm,algpseudocode}
\usepackage{algorithmicx}

\usepackage{wrapfig}
\usepackage{pifont}
\usepackage{bbding}
\usepackage{xspace}
\usepackage{multicol}

\newcommand{\model}{\textsc{DiCE}}
\newcommand{\cmark}{\textcolor{green!60!black}{\ding{51}}}
\newcommand{\xmark}{\textcolor{red}{\ding{55}}}

\makeatletter
\DeclareRobustCommand\onedot{\futurelet\@let@token\@onedot}
\def\@onedot{\ifx\@let@token.\else.\null\fi\xspace}

\makeatother

\title{\underline{Di}vide-then-Diagnose: Weaving Clinician-Inspired Contexts for Ultra-Long \underline{C}apsule \underline{E}ndoscopy Videos}

\author[1]{Bowen Liu}
\author[2]{Li Yang}
\author[1]{Shanshan Song}
\author[2]{Mingyu Tang}
\author[2]{Zhifang Gao}
\author[1]{Qifeng Chen}
\author[1]{Yangqiu Song}
\author[2]{Huimin Chen}
\author[1]{Xiaomeng Li}

\affiliation[1]{The Hong Kong University of Science and Technology}
\affiliation[2]{Renji Hospital, Shanghai Jiao Tong University School of Medicine}

\correspondence{Xiaomeng Li, Huimin Chen}

\abstract{Capsule endoscopy (CE) enables non-invasive gastrointestinal screening, but current CE research remains largely limited to frame-level classification and detection, leaving video-level analysis underexplored. To bridge this gap, we introduce and formally define a new task, \emph{diagnosis-driven CE video summarization}, which requires extracting  key evidence frames that covers clinically meaningful findings and making accurate diagnoses from those evidence frames. This setting is challenging because diagnostically relevant events are extremely sparse and can be overwhelmed by tens of thousands of redundant normal frames, while individual observations are often ambiguous due to motion blur, debris, specular highlights, and rapid viewpoint changes. To facilitate research in this direction, we introduce VideoCAP, the first CE dataset with diagnosis-driven annotations derived from real clinical reports. VideoCAP comprises 240 full-length videos and provides realistic supervision for both key evidence frame extraction and diagnosis. To address this task, we further propose \model, a clinician-inspired framework that mirrors the standard CE reading workflow. \model~first performs efficient candidate screening over the raw video, then uses a Context Weaver to organize candidates into coherent diagnostic contexts that preserve distinct lesion events, and an Evidence Converger to aggregate multi-frame evidence within each context into robust clip-level judgments. Experiments show that \model~consistently outperforms state-of-the-art methods, producing concise and clinically reliable diagnostic summaries. These results highlight diagnosis-driven contextual reasoning as a promising paradigm for ultra-long CE video summarization.}

\begin{document}

\maketitle

\newcommand{\tabincell}[2]{\begin{tabular}{@{}#1@{}}#2\end{tabular}}

\section{Introduction}
\begin{figure}[t]
    \centering
    \includegraphics[width=0.8\linewidth]{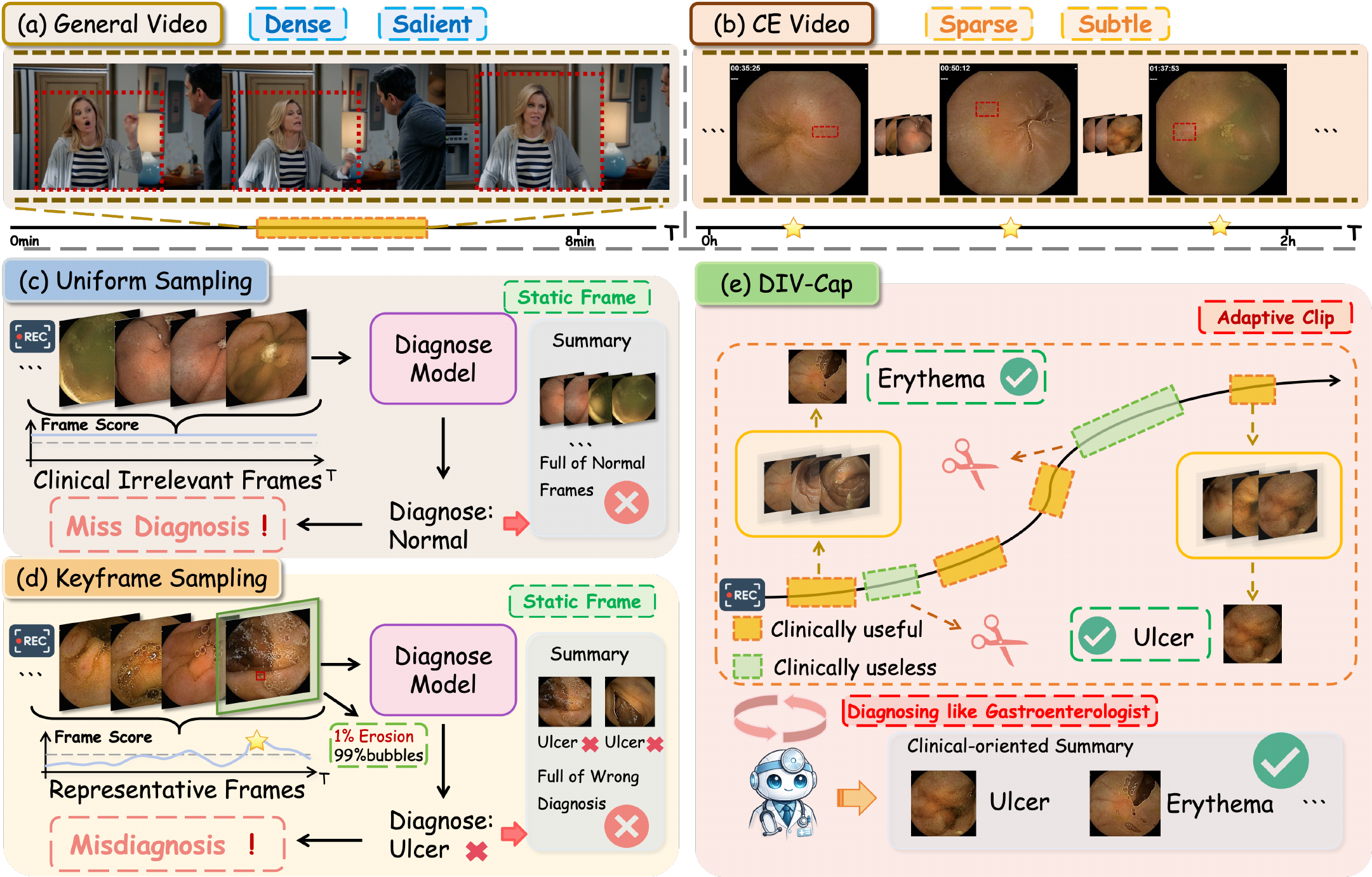}
    \caption{
(a) \textbf{General videos} typically contain task-relevant evidence that is temporally dense and visually salient, as reflected by the larger red box and denser timeline;
(b) \textbf{CE videos} instead contain diagnostic evidence that is temporally sparse and visually subtle, as reflected by the smaller red box and sparser timeline;
(c) \textbf{Uniform sampling} selects clinically irrelevant frames due to the sparse nature of lesions, leading to missed diagnoses;
(d) \textbf{Keyframe sampling} captures unsatisfactory representative frames (e.g., bubbles), resulting in misdiagnosis;
(e) \textbf{\model~(Ours)} adaptively extracts context-aware evidence clips from the raw video, rather than selecting a fixed frame budget, to generate clinically oriented summaries and support correct diagnosis.}
    \label{fig:intro}
   \vspace{-5mm}
\end{figure}

Capsule endoscopy (CE)\cite{andrade_ai-assisted_2025,habe_capsule_2024,panchananam_capsule_2024,xie_development_2022} enables non-invasive gastrointestinal screening and has become an essential tool for large-scale outpatient diagnosis. However, a single examination generates an ultra-long video stream lasting 8--12 hours\cite{xie_development_2022}, and reviewing such data remains time-consuming and cognitively demanding: even experienced clinicians typically spend over an hour per study. This heavy workload limits clinical scalability and increases the risk of missed lesions~\cite{lancetdh,xie_development_2022}. Existing automated CE methods mainly focus on frame-level classification~\cite{classify2,classify3,classify5,classify6,classify7,classify8} and detection~\cite{detection,detection1,detection2,detection3,detection5,detection6} on curated image datasets, where diagnostically relevant frames have already been manually isolated into clean, high-quality examples and the goal is to recognize whether an individual frame contains a lesion. Clinical CE reading, however, is a full-video diagnostic process: physicians inspect the entire examination video and retain only the lesion-related evidence that supports the final report. This mismatch leaves the clinically realistic problem of full-video CE summarization largely unexplored. To bridge this gap, we introduce a new task, \emph{diagnosis-driven CE video summarization}, which aims to extract concise yet diagnosis-supporting visual evidence from raw CE videos. 

One simple method for CE video summarization is to apply existing frame-level CE models to every frame in the raw video and then aggregate the predictions. However, this strategy achieves limited clinical utility: deployment studies report that after frame-level analysis, only 8\% of selected images contained significant lesions~\cite{haslach2023reading}, and physician review time still remained more than 1 hour~\cite{oh2024reading}. In real CE examinations, diagnostically relevant frames are overwhelmed by vast numbers of normal views, while capsule motion, motion blur, luminal debris, specular highlights, and rapidly changing viewpoints make isolated frames noisy and diagnostically ambiguous. Under this regime, frame-level predictions on raw videos become unstable and redundant, and simple post-hoc aggregation still fails to consolidate uncertain observations into robust diagnostic judgments. As shown in Figure~\ref{fig:uncertainty}, roughly half of nearby keyframes from strong baselines receive conflicting diagnoses. This exposes the first challenge of diagnosis-driven CE summarization: \textbf{high uncertainty of isolated frames}, which makes simple frame-by-frame inference and aggregation insufficient for robust diagnosis.

Another feasible route is to adapt natural long-video understanding methods to CE summarization~\cite{qwen2vl,qwen2.5vl,internvl,internvl3,internvl3.5,AKS,tang_tspo_2025,vilamp,videoagent,longvideoagent}. The simplest adaptation is uniform sampling~\cite{qwen2vl,qwen2.5vl,internvl,internvl3,internvl3.5}, which sparsely samples frames from the full video to keep long-horizon reasoning tractable. Later methods replace uniform sampling with keyframe selection~\cite{AKS,tang_tspo_2025,vilamp,videoagent,longvideoagent}, aiming to retain more salient or representative frames for downstream reasoning. In practice, however, both strategies are poorly matched to CE. Figure~\ref{fig:intro}a--b highlights the core mismatch: in many natural-video settings, informative events are temporally dense and visually salient enough to survive representative sampling, whereas CE lesion cues are sparse, subtle, short-lived, and often embedded in redundant normal or artifact-heavy views. Figure~\ref{fig:intro}c--d further illustrates the resulting failure modes: uniform sampling tends to select clinically irrelevant frames because lesions are rare, whereas keyframe selection may favor visually representative yet diagnostically unsatisfactory frames such as bubbles. A standard CE video may comprise as many as 100k frames, while fewer than ten may be diagnostically relevant for final clinical decision-making~\cite{ibd1,ibd2}. This exposes the second challenge of diagnosis-driven CE summarization: \textbf{extreme spatio-temporal sparsity of clinically relevant events}. Taken together, these two challenges suggest that CE summarization should be formulated not as selecting representative frames, but as extracting diagnostically sufficient evidence from sparse and uncertain lesion events.

The clinical reading process offers a natural blueprint for addressing these challenges: clinicians first localize potentially relevant events throughout the video and then integrate evidence across temporally adjacent views before committing to a judgment. Inspired by this workflow, we propose \model~(\textbf{Di}vide-then-Diagnose for \textbf{C}apsule \textbf{E}ndoscopy), the first diagnosis-driven framework that explicitly shifts CE analysis from frame-level analysis to context-based reasoning(Figure~\ref{fig:intro}e). Rather than treating a CE video as an undifferentiated frame stream, \model~first \emph{divides} the video into coherent diagnostic contexts centered on distinct lesion, and then \emph{diagnoses} each context by converging multi-frame evidence into a robust clinical judgment.

Concretely, \model~realizes this design through two key reasoning stages, supported by an efficient preprocessing step. A lightweight \emph{Selector} first performs high-recall screening over the raw video stream, reducing tens of thousands of frames to a manageable candidate set. On top of this candidate set, the \emph{Context Weaver} organizes frames into temporally and visually coherent diagnostic contexts so that rare lesion events can be preserved as distinct evidence units rather than being overwhelmed by redundant normal views. The \emph{Evidence Converger} then performs multi-frame synthesis within each context, aggregating noisy observations into a single robust diagnostic judgment.

To support model development and evaluation, we introduce VideoCAP, the first diagnosis-driven CE dataset derived from real clinical reports and comprising 240 full-length patient videos with report-derived annotations. Experiments on VideoCAP show that \model~outperforms state-of-the-art methods, supporting the value of diagnosis-driven contextual reasoning for ultra-long CE video summarization.

Our key contributions are as follows:
\begin{itemize}
\item[\ding{182}] \textbf{New Task formulation.} We are the first to propose and formally
define the novel task of diagnosis-driven
CE video summarization.
\item[\ding{183}] \textbf{Novel Clinician-inspired framework.} We propose \model, the first diagnosis-driven framework that explicitly shifts CE analysis from frame-level analysis to contextual reasoning for summarization of ultra-long CE videos.
\item[\ding{184}] \textbf{New Dataset and benchmark.} We introduce VideoCAP, the first diagnosis-driven CE dataset derived from real clinical reports, comprising 240 full-length videos with report-derived annotations.
\item[\ding{185}] \textbf{Experimental validation.} Extensive experiments show that \model~outperforms state-of-the-art methods, validating diagnosis-driven contextual reasoning as an effective paradigm for ultra-long CE video summarization.

\end{itemize}

   \section{Related Work}

  \subsection{Long Video Analysis}

  Long-video methods typically reduce computation through sparse sampling or frame selection. Strong video understanding models such as QwenVL and the InternVL family rely on sparse visual inputs with efficient long-context modeling~\cite{qwen2vl,qwen2.5vl,internvl,internvl3,internvl3.5}, while keyframe sampling methods first select representative or query-relevant frames before downstream reasoning~\cite{AKS,tang_tspo_2025,vilamp,videoagent,longvideoagent}. In contrast, diagnosis-driven CE summarization requires localize rare lesion events and aggregating evidence within local contexts, rather than preserving representative content under a fixed frame budget.

  \subsection{AI for Capsule Endoscopy Analysis and Datasets}

  Most CE studies target frame-level lesion classification or detection~\cite{classify2,classify3,classify5,classify6,classify7,classify8,detection,detection1,detection2,detection3,detection5,detection6}. Public datasets such as Kvasir-Capsule~\cite{smedsrud2021kvasir}, the Rhode Island corpus~\cite{charoen2022rhode}, and Galar~\cite{lefloch2025galar} have advanced this line, but they mainly provide curated image-level supervision rather than full-video, diagnosis-driven annotations. Beyond frame recognition, prior CE work has explored anomaly scoring~\cite{mohammed_psdevcem_2020}, keyframe extraction~\cite{sushma_summarization_2021}, abnormal-frame pre-filtering~\cite{morera_reduction_2022,pinto_reducing_2025}, temporal modeling~\cite{oh_video_2023}, transit-time estimation~\cite{nam_deep_2024}, and robustness under distribution shift~\cite{tan_endoood_2024,Liu_2025_sugar}. However, these methods and datasets still provide limited support for diagnosis-driven full-video summarization, where the model must localize sparse evidence and consolidate it into reliable judgments. Our work fills this gap with VideoCAP and a contextual reasoning framework designed for full-length CE videos.

\setlength{\tabcolsep}{3pt}

\section{Dataset and Benchmark}
\label{sec:dataset}

\begin{table*}[t]
\renewcommand{\arraystretch}{1.3}
\centering
\caption{Comparison with existing capsule endoscopy datasets. ``--'' indicates unavailable information. VideoCAP is the first CE dataset with diagnosis-driven annotations derived from real clinical reports, where each annotation pairs a lesion label with a diagnostic keyframe timestamp.}
\label{table:dataset_comparison}
\begin{adjustbox}{width=0.9\linewidth}
\begin{tabular}{lccccccc}
\toprule
\textbf{Dataset}
& \textbf{Video}
& \textbf{Lesion Anno.}
& \textbf{\#Videos}
& \textbf{Resolution}
& \textbf{\#Frames/Images}
& \textbf{Diag. Report}
& \textbf{Diag. Timestamp} \\
\midrule
Rhode Island \cite{charoen2022rhode}
                & \xmark & \xmark & --
                & 320$\times$320
                & 5,247,588 & \xmark & \xmark \\
AI-KODA \cite{handa2024aikoda}
                & \xmark & \cmark & --
                & 320$\times$320
                & 2,173     & \xmark & \xmark \\
SEE-AI \cite{seeai2022}
                & \xmark & \cmark & --
                & --
                & 18,481    & \xmark & \xmark \\
Kvasir-Capsule \cite{smedsrud2021kvasir}
                & \cmark & \cmark & 43
                & 256$\times$256 -- 512$\times$512
                & 1,955,675 & \xmark & \xmark \\
KID \cite{koulaouzidis2017kid}
                & \cmark & \cmark & 47
                & 360$\times$360
                & 2,500     & \xmark & \xmark \\
Galar \cite{lefloch2025galar}
                & \cmark & \cmark & 80
                & 336$\times$336 -- 512$\times$512
                & 3,513,539 & \xmark & \xmark \\
\rowcolor[HTML]{CFF2DF}
\textbf{VideoCAP (Ours)}
                & \textbf{\cmark} & \textbf{\cmark} & \textbf{240}
                & \textbf{576$\times$576}
                & \textbf{7,245,249} & \textbf{\cmark} & \textbf{\cmark} \\
\bottomrule
\end{tabular}
\end{adjustbox}
\vspace{1mm}

\raggedright\footnotesize{
\textbf{Diag. Report}: annotations derived from clinical diagnostic reports, capturing only lesions that contributed to the final patient diagnosis. \textbf{Diag. Timestamp}: timestamp of the diagnostic keyframe corresponding to each clinically reported finding.
}
\end{table*}

Existing CE datasets (Table~\ref{table:dataset_comparison}) are predominantly image-centric: they curate selected images into standalone collections and label every visible abnormality, regardless of whether a finding actually contributed to the patient's final diagnosis. As a result, they provide limited support for evaluating diagnosis-driven CE summarization, which requires full-length video context, timestamped diagnostic keyframes, and patient-level clinical grounding. To bridge this gap, we introduce VideoCAP, a diagnosis-driven dataset comprising 240 full-length CE videos collected from two clinical centers of Shanghai Renji Hospital.\\
\begin{wrapfigure}{r}{0.45\linewidth}
  \vspace{-10pt}
  \centering
  \includegraphics[width=\linewidth]{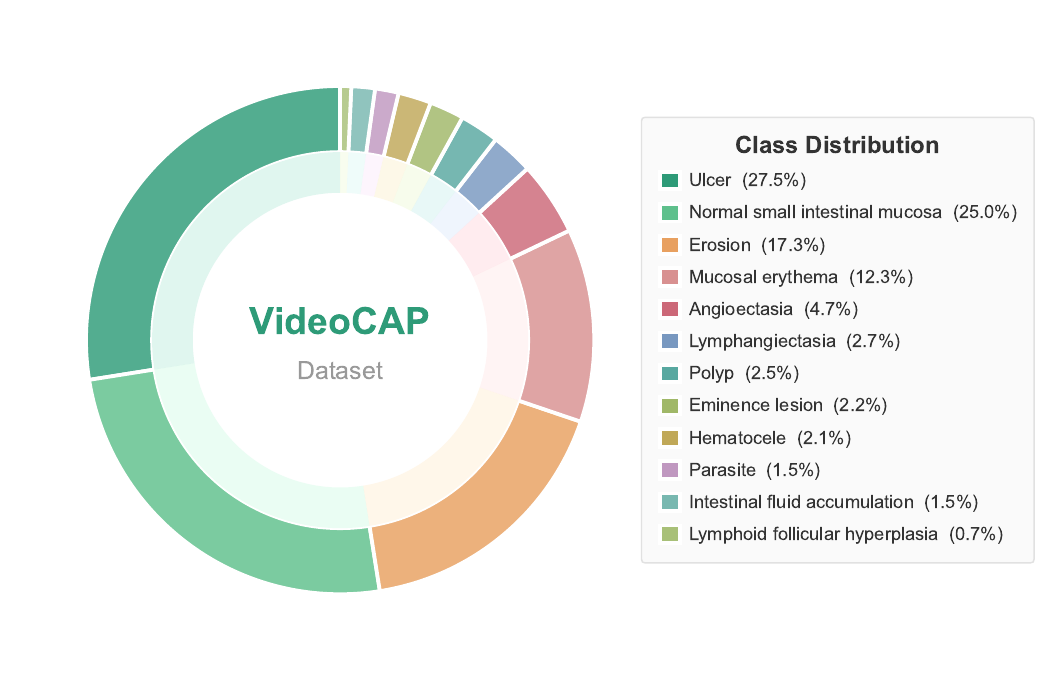}
  \caption{Dataset statistics.}
  \label{fig:data}
  \vspace{-30pt}
\end{wrapfigure}
\subsection{Data Collection and Annotation}
\label{subsec:data_collection}

All videos were captured during routine clinical examinations at a native resolution of $576 \times 576$. The dataset preserves real-world CE variability, including extremely sparse lesion occurrences, frequent motion blur, luminal debris, specular highlights, and rapid viewpoint changes. We split the data into 160/40/40 videos for training, validation, and testing at the patient level to prevent data leakage.\\

\paragraph{Annotation protocol.}
Unlike prior datasets that retrospectively label curated image collections---often including all visible abnormalities regardless of diagnostic significance---our annotations are derived directly from clinical diagnostic reports issued during routine patient care. Each report entry corresponds to a clinically relevant finding that directly contributed to the patient's final diagnosis, ensuring that the benchmark captures clinically actionable evidence rather than exhaustive image-level cataloging. Every report was further reviewed and verified by three senior gastroenterologists to establish a gold-standard reference. The resulting annotations are organized as \emph{diagnosis-driven annotations}: each annotated finding records the lesion label together with the timestamp of its corresponding diagnostic keyframe in the full examination video. Because these diagnoses are made in the context of the overall patient examination rather than as isolated image labels, they better reflect clinically meaningful diagnostic decisions. The annotation taxonomy covers 12 clinically established categories---ulcer, erosion, angioectasia, mucosal erythema, eminence lesion, hematocele, lymphangiectasia, lymphoid follicular hyperplasia, polyp, parasite, intestinal fluid accumulation, and normal small intestinal mucosa---aligned with standard CE reporting guidelines~\cite{pennazio2023small}. This diagnosis-driven annotation scheme supports lesion-level, keyframe-level, and patient-level evaluation.

\subsection{Evaluation Protocol}
\label{subsec:eval_protocol}

We evaluate all methods under a unified protocol at three complementary granularities: lesion-level, keyframe-level, and patient-level. All metrics are computed after greedy one-to-one temporal deduplication against annotated diagnostic keyframes so that repeated detections of the same clinically reported finding are not over-counted.

\paragraph{Matching rule.}
Following established CE clinical studies~\cite{lancetdh}, a selected frame is considered matched to an annotated diagnostic keyframe if it falls within a $\pm 300$\,s temporal window around the keyframe timestamp. To penalize temporally proximate but semantically inconsistent predictions, two selected frames within 20\,s of each other that carry different predicted labels are both treated as incorrect. When multiple selected frames match the same clinically reported finding, only the frame with the smallest time error is retained as the true match, and the rest count toward redundancy.

\paragraph{Lesion-level metrics.}
\emph{Lesion Detection Rate} (LDR) measures the proportion of clinically reported findings that are both temporally matched and assigned the correct lesion category. \emph{Sensitivity} measures the proportion of clinically reported findings that are hit by at least one temporally matched selected frame, regardless of the predicted lesion category, and therefore captures coverage of clinically reported findings. \emph{Specificity} measures the fraction of unique predicted lesion findings that correspond to a clinically reported finding after deduplication, reflecting how often the reported findings are clinically valid.

\paragraph{Keyframe-level metrics.}
\emph{Time Error} = \(|t_{\text{frame}} - t_{\text{keyframe}}|\) measures temporal localization in seconds relative to the annotated diagnostic keyframe timestamp for correctly matched findings. \emph{Redundancy}  = \(\frac{\#\text{selected} - \#\text{matched}}{\#\text{selected}}\) captures the fraction of selected frames that contribute no new lesion information after deduplication.

\paragraph{Patient-level metrics.}
\emph{Diagnostic Yield} (DY) is the fraction of patients for whom \emph{all} clinically reported findings are successfully detected. \emph{Per-Patient Detection Rate} (PDR) is the fraction of patients with at least one correctly detected lesion.

\section{Method}
\label{sec:method}
\begin{figure*}[t]
    \centering
    \includegraphics[width=0.8\linewidth]{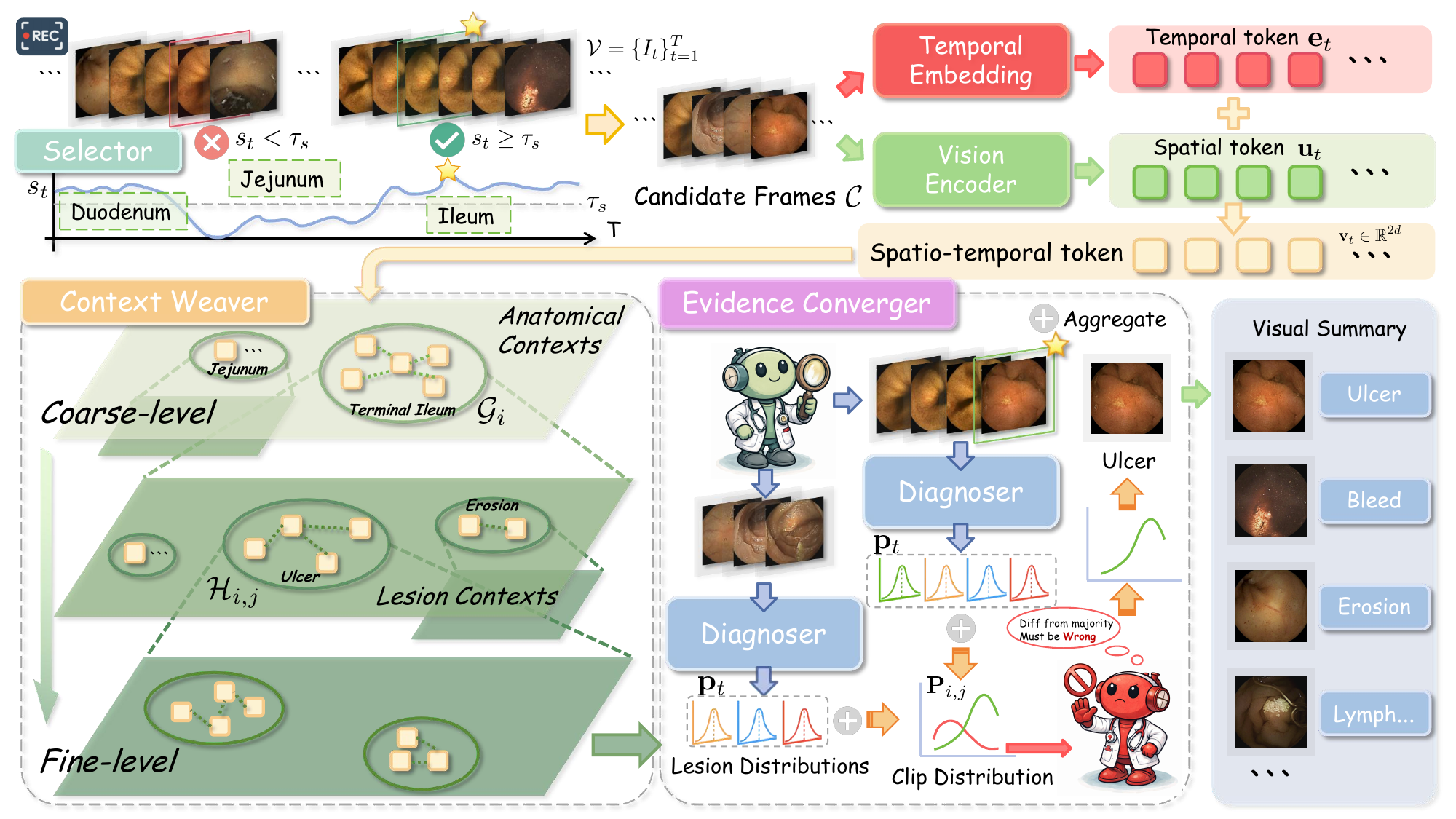}
    \caption{Overview of \model. A \textbf{Selector} first filters the raw video into a high-recall candidate pool. Each retained frame is encoded as a spatio-temporal token combining appearance and temporal position. The \textbf{Context Weaver} then constructs a two-level hierarchy of coarse anatomical contexts and fine lesion contexts. Finally, the \textbf{Evidence Converger} aggregates frame-level predictions within each lesion context into a stable context-level diagnosis and removes inconsistent frames, yielding a concise diagnostic summary.}

\vspace{-5mm}
    \label{fig:overview}
\end{figure*}
We present \model, a diagnosis-driven framework for summarizing long CE videos into concise and diagnostically meaningful visual summaries. The framework follows a coarse-to-fine reasoning pipeline. After an initial high-recall screening stage (Selector $\mathcal{S}$) filters the raw video stream into diagnostically relevant candidates, \model~performs contextual reasoning through two core modules: a Context Weaver $\mathcal{W}$ that organizes candidates into hierarchical diagnostic contexts, and an Evidence Converger $\mathcal{E}$ that aggregates multi-frame evidence within each context into robust context-level diagnoses. To enable the Context Weaver to jointly reason about appearance and temporal-anatomical position, we first construct spatio-temporal tokens that fuse visual features with temporal encodings. We next introduce the notation and then detail the Selector, Context Weaver, and Evidence Converger in turn.

\subsection{Notation}

A CE video is denoted $\mathcal{V}=\{I_t\}_{t=1}^{T}$. Given $\mathcal{V}$, our goal is to produce a compact visual summary
\begin{equation}\label{eq:report_def}
    \mathcal{R} = \bigl[(\tilde{I}_{i,j},\;\hat{y}_{i,j})\bigr]\,,
\end{equation}
where each entry corresponds to one retained lesion context: $\tilde{I}_{i,j}$ is the representative keyframe of the $j$-th fine-grained lesion context inside the $i$-th coarse anatomical context, and $\hat{y}_{i,j}$ is the corresponding context-level lesion prediction. Here, $\mathcal{R}$ denotes the visual summary. Throughout the section, $t$ indexes video frames, $i$ indexes coarse anatomical contexts, and $j$ indexes fine lesion contexts within each coarse context. Ideally, the summary should cover distinct lesion findings while avoiding redundant or spurious findings.

\subsection{Selector $\mathcal{S}$}

The Selector $\mathcal{S}$ serves as a high-recall preprocessing stage that performs efficient binary screening over the raw video stream frame by frame:
\begin{equation}\label{eq:selector}
    s_t = h_\psi\!\bigl(f_\theta(I_t)\bigr) \in [0,1]\,,
\end{equation}
where $f_\theta$ is a frozen vision backbone producing a visual feature $\mathbf{u}_t = f_\theta(I_t) \in \mathbb{R}^{d}$, and $h_\psi$ is a lightweight MLP head trained with binary cross-entropy to distinguish diagnostically relevant frames from normal ones. Frames with $s_t \ge \tau_s$ are retained as the candidate set:
\begin{equation}\label{eq:candidates}
    \mathcal{C} = \{I_t \mid s_t \ge \tau_s\}\,,
\end{equation}
substantially reducing the search space while preserving frames that may contribute to the final diagnosis.

\subsection{Spatio-Temporal Tokenization}
\label{subsec:st_token}

Frame-level analysis ignores the temporal continuity of CE examinations: nearby frames often show the same region from slightly different viewpoints. To encode this continuity, we construct a \emph{spatio-temporal token} for each candidate frame by fusing its visual feature with a temporal signal. Specifically, we compute a sinusoidal temporal embedding $\mathbf{e}_t \in \mathbb{R}^{d}$ from the frame index $t$ and form:
\begin{equation}\label{eq:token}
    \mathbf{v}_t = \!\bigl(\mathbf{u}_t \oplus \mathbf{e}_t\bigr) \in \mathbb{R}^{2d}\,,
\end{equation}
where $\oplus$ denotes concatenation. The token $\mathbf{v}_t$ jointly encodes frame content and examination position.

\subsection{Context Weaver $\mathcal{W}$}

Existing keyframe selection methods evaluate frames in isolation and ignore their position along the gastrointestinal tract. This can over-select visually salient regions, miss subtle lesions elsewhere, and merge nearby but distinct findings. The Context Weaver $\mathcal{W}$ addresses this by constructing a hierarchy of \emph{diagnostic contexts} from coarse anatomical coverage to fine lesion-level specificity.

\paragraph{Hierarchical context construction.}
The Context Weaver takes the candidate tokens $\{\mathbf{v}_t \mid I_t \in \mathcal{C}\}$ and organizes them into a two-level contextual hierarchy:
\begin{equation}\label{eq:decompose}
    \underbrace{\mathcal{C}}_{\text{candidates}}
    \;\xrightarrow{\;\mathcal{W}_{\text{coarse}}\;}\;
    \underbrace{\{\mathcal{G}_i\}_{i=1}^{N_c}}_{\text{anatomical contexts}}
    \;\xrightarrow{\;\mathcal{W}_{\text{fine}}\;}\;
    \underbrace{\{\mathcal{H}_{i,j}\}_{j=1}^{M_i}}_{\text{lesion contexts}}\,,
\end{equation}
where $\mathcal{G}_i$ denotes a coarse context associated with a contiguous portion of the examination trajectory, and $\mathcal{H}_{i,j}$ denotes a fine-grained lesion context. At both levels, grouping is driven by the joint temporal-visual affinity encoded in $\mathbf{v}_t$, but at different resolutions. Specifically, $\mathcal{W}_{\text{coarse}}$ assembles candidate tokens into $N_c$ progression-aware groups according to their joint temporal-visual compatibility, and $\mathcal{W}_{\text{fine}}$ further refines each coarse group into $M_i$ lesion-focused contexts using the same compatibility at a finer granularity. Thus, the coarse stage distributes attention across the examination, whereas the fine stage separates nearby lesion hypotheses before diagnosis.

\paragraph{Anatomical context anchoring (coarse level).}
Although capsule motion is irregular, CE still follows a coarse anatomical progression, so temporal position is a useful proxy. The coarse stage therefore groups tokens with compatible temporal position and visual appearance into broad contiguous contexts without requiring explicit organ labels. Intuitively, different coarse contexts may correspond to different phases of the examination, such as earlier small-bowel views around the proximal jejunum versus later views closer to the terminal ileum. Rather than repeatedly selecting frames from a few salient regions, the $N_c$ anchors encourage broader coverage of distinct events across the bowel.

\paragraph{Lesion context refinement (fine level).}
Within one coarse context $\mathcal{G}_i$, multiple phenomena may still co-exist: repeated glimpses of one lesion, nearby but distinct lesions, and occasional irrelevant normal or low-quality frames. The fine stage therefore decomposes each $\mathcal{G}_i$ into $M_i$ lesion contexts $\{\mathcal{H}_{i,j}\}$ by applying finer-grained grouping within the coarse context, so that visually and temporally coherent observations supporting different lesion hypotheses can be separated before diagnosis. To describe the intended structure of a well-formed lesion context, let $y_t^{*}$ denote the lesion label of frame $I_t$, and let $y_{i,j}^{*}$ denote the latent dominant label associated with context $\mathcal{H}_{i,j}$. We use the following dominant-label prior:
\begin{equation}\label{eq:invariant}
    y_{i,j}^{*} = \operatorname*{mode}_{I_t \in \mathcal{H}_{i,j}} y_t^{*}\,,
\end{equation}
\emph{i.e.}, a well-formed lesion context should be dominated by one underlying finding, although a small number of irrelevant or corrupted frames may remain. This prior is conceptual rather than explicitly supervised, and is used to motivate the subsequent context-level evidence fusion and refinement.

\subsection{Evidence Converger $\mathcal{E}$}

Motivated by the dominant-label prior in Eq.~\ref{eq:invariant}, the Evidence Converger $\mathcal{E}$ performs diagnosis at the lesion-context level: rather than trusting any single frame, it converts each lesion context into a stable diagnostic judgment through three stages---multi-frame evidence fusion, intra-context refinement, and inter-context pruning.

\begin{table}[t]
\renewcommand{\arraystretch}{1.3}
\centering
\caption{Zero-shot evaluation results on capsule endoscopy lesion detection and keyframe quality. Size indicates model parameters. Frames denotes the frame sampling strategy or keyframe selection budget. }
\label{table:zero_shot_main}
\begin{adjustbox}{width=\linewidth}
\begin{tabular}{l c c c c c c c}
\toprule
\multicolumn{1}{c}{} 
& \multicolumn{1}{c}{} 
& \multicolumn{1}{c}{} 
& \multicolumn{3}{c}{\textbf{Lesion-Level}} 
& \multicolumn{2}{c}{\textbf{Frame-level}} \\
\cmidrule(lr){4-6} \cmidrule(l){7-8}
\textbf{Models} 
& \textbf{Size} 
& \textbf{Frames}
& \textbf{Detection Rate (\%)} 
& \textbf{Sensitivity (\%)} 
& \textbf{Specificity (\%)} 
& \textbf{Time Error (sec)} 
& \textbf{Redundancy (\%)} \\
\midrule
\midrule
\rowcolor[HTML]{DDBBF2}
\multicolumn{8}{c}{\textit{\textbf{Colon Specific}}} \\ 
\midrule

ColonX\cite{ji_colon-x_2025}     & 3B & 32 & 2.94  & 72.06 & 20.58 & 49.91 & 79.42 \\
ColonGPT\cite{colongpt}   & 2B & 32 & 9.56  & 69.85 & 21.59 & 54.11 & 78.41 \\
\midrule

\rowcolor[HTML]{DDBBF2}
\multicolumn{8}{c}{\textit{\textbf{General Medical}}} \\ 
\midrule

MedGemma\cite{medgemma}   & 4B & 32 & 9.56  & 62.50 & 21.59 & 64.74 & 78.41 \\
HuatuoGPT\cite{huatuogptvision}  & 7B & 32 & 8.09  & 61.03 & 20.58 & 55.00 & 79.42 \\
LingShu\cite{xu2025lingshu}    & 7B & 32 & 17.65 & 61.76 & 20.36 & 66.21 & 79.64 \\
\bottomrule
\end{tabular}
\end{adjustbox}
\end{table}

\begin{table}[t]
\renewcommand{\arraystretch}{1.3}
\small
\centering
\caption{Full training evaluation results on capsule endoscopy keyframe detection and diagnostic performance. Size indicates model parameters. Frames denotes the frame sampling strategy or keyframe selection budget. Best results  are in \textbf{bold}, second best results are \underline{underlined}.}
\label{table:main_results}
\begin{adjustbox}{width=\linewidth}
\small
\begin{tabular}{>{\footnotesize}l c c c c c c c c c}
\toprule
\multicolumn{1}{c}{} & \multicolumn{1}{c}{} & \multicolumn{1}{c}{} & \multicolumn{3}{c}{\textbf{Lesion-Level}} & \multicolumn{2}{c}{\textbf{Frame-level}} & \multicolumn{2}{c}{\textbf{Patient-Level}} \\
\cmidrule(lr){4-6} \cmidrule(l){7-8} \cmidrule(l){9-10}
\multicolumn{1}{c}{\multirow{-2}{*}{\textbf{Models}}} & \multicolumn{1}{c}{\multirow{-2}{*}{\textbf{Size}}} & \multirow{-2}{*}{\textbf{Frames}} & \textbf{Detection Rate} & \textbf{Sensitivity} & \textbf{Specificity} & \textbf{Time Error} & \textbf{Redundancy} & \textbf{Diagnostic Yield} & \textbf{Detection Rate} \\
 & & & \textbf{(\%)} & \textbf{(\%)} & \textbf{(\%)} & \textbf{(sec)} & \textbf{(\%)} & \textbf{(\%)} & \textbf{(\%)} \\
\midrule
\midrule
\rowcolor[HTML]{DDBBF2}
\multicolumn{10}{c}{\textit{\textbf{General}}} \\ \midrule
Qwen3-VL + AKS & 8B & 32 & 18.38 & 48.53 & 19.19 & 78.42 & 91.73 & 2.50 & 32.50 \\
Qwen3-VL + ViLAMP & 8B & 32 & 21.32 & 65.44 & 18.83 & 70.82 & 88.69 & 2.50 & 35.00 \\
\midrule
InternVL3.5 + AKS & 8B & 32 & 19.85 & 51.47 & \textbf{23.55} & 67.77 & 83.45 & 5.00 & 32.50 \\
InternVL3.5 + ViLAMP & 8B & 32 & 27.94 & 66.18 & 22.62 & 66.28 & \underline{78.86} & 7.50 & 40.00 \\
\midrule
Dinov3 + AKS & 0.2B & 32 & 33.82 & 50.00 & 22.88 & 69.13 & 84.78 & \underline{15.00} & 60.00 \\
Dinov3 + ViLAMP & 0.2B & 32 & \underline{35.29} & 61.76 & \underline{23.18} & 69.69 & 83.26 & 12.50 & 60.00 \\
\midrule
\midrule
\rowcolor[HTML]{DDBBF2}
\multicolumn{10}{c}{\textit{\textbf{Medical}}} \\ \midrule
Colongpt + AKS & 2B & 32 & 25.00 & 50.74 & 23.05 & 78.16 & 89.41 & 10.00 & 45.00 \\
Colongpt + ViLAMP & 2B & 32 & 34.56 & 63.97 & 21.86 & 65.21 & 85.07 & 12.50 & 57.50 \\
\midrule

Lingshu + AKS & 7B & 32 & 27.21 & 50.74 & 21.86 & 64.54 & 87.48 & 10.00 & 52.50 \\
Lingshu + ViLAMP & 7B & 32 & \underline{35.29} & \underline{67.65} & 21.80 & \underline{59.91} & 82.32 & 7.50 & \underline{65.00} \\
\midrule
\midrule
\rowcolor[HTML]{CFF2DF}
\multicolumn{1}{l}{\model (Ours)} & 0.2B & 29.98 & \textbf{44.12} & \textbf{85.29} & 19.30 & \textbf{54.86} & \textbf{77.81} & \textbf{20.00} & \textbf{67.50} \\
\bottomrule
\end{tabular}
\end{adjustbox}
\end{table}

\paragraph{Multi-frame evidence fusion.}
A lesion diagnoser $g_\phi$ maps each frame to a categorical distribution over $K$ lesion types:
\begin{equation}\label{eq:frame_pred}
    \mathbf{p}_t = g_\phi(I_t) \in \Delta^{K-1}\,.
\end{equation}
We refer to $\mathbf{p}_t$ as the \emph{lesion distribution} of frame $I_t$. To obtain a provisional context-level judgment, we aggregate all frame-level lesion distributions within a lesion context into an unnormalized \emph{context evidence vector}:
\begin{equation}\label{eq:fuse}
    \mathbf{P}_{i,j}
    = \sum_{I_t \in \mathcal{H}_{i,j}} \mathbf{p}_t\,,
    \qquad
    \hat{y}^{(0)}_{i,j} = \arg\max_k\;\mathbf{P}_{i,j}[k]\,.
\end{equation}
When a context is dominated by one lesion hypothesis, summation strengthens consistent evidence while suppressing sporadic mispredictions caused by transient blur, debris, or suboptimal viewing angles. This converts a set of uncertain frame-level predictions into a single, more robust provisional diagnosis.

\paragraph{Intra-context coherence refinement.}
Aggregation handles random noise effectively, but it can still be biased by frames that are \emph{systematically} corrupted (\emph{e.g.}, a debris-occluded view that persistently activates a wrong category). We again leverage Eq.~\ref{eq:invariant}: in a well-formed lesion context, most frames should support the same dominant lesion hypothesis, so frames that contradict the context consensus are likely unreliable. We first keep the frames that agree with the provisional context-level decision,
\begin{align}
    \widetilde{\mathcal{H}}_{i,j}
    &= \bigl\{\,I_t \in \mathcal{H}_{i,j}
      \;\bigm|\; \mathbf{p}_t[\hat{y}^{(0)}_{i,j}] \ge \tau_{\text{agree}}\,\bigr\}\,,
    \label{eq:retain_tmp}\\[4pt]
    \mathcal{H}_{i,j}^{\,\text{ret}}
    &=
    \begin{cases}
        \widetilde{\mathcal{H}}_{i,j}, & \widetilde{\mathcal{H}}_{i,j} \neq \emptyset,\\
        \mathcal{H}_{i,j}, & \text{otherwise},
    \end{cases}
    \label{eq:retain}\\[4pt]
    \mathbf{P}_{i,j}^{\,\text{ref}}
    &= \sum_{I_t \in \mathcal{H}_{i,j}^{\,\text{ret}}} \mathbf{p}_t\,.
    \label{eq:refine}
\end{align}
We then recompute the final context-level prediction as
\begin{equation}\label{eq:refine_pred}
    \hat{y}_{i,j} = \arg\max_k\;\mathbf{P}_{i,j}^{\,\text{ref}}[k]\,.
\end{equation}
Here $\tau_{\text{agree}}$ is a consistency threshold. If thresholding removes every frame, we fall back to the original context to avoid empty refined contexts. This self-consistency check sharpens the final context-level prediction by filtering out diagnostically incoherent frames.

\paragraph{Inter-context pruning.}
While intra-context refinement improves the quality of each individual lesion context, the final set may still contain non-informative or unreliable entries. We therefore discard contexts that remain normal or low-confidence after refinement:
\begin{equation}\label{eq:prune}
    \text{Discard } \mathcal{H}_{i,j}^{\,\text{ret}} \quad \text{if} \quad
    \hat{y}_{i,j} = \texttt{normal}
    \;\;\lor\;\;
    c_{i,j} < \tau_{\min}\,,
\end{equation}
where $c_{i,j} = \max_k\, \bar{\mathbf{P}}_{i,j}^{\,\text{ref}}[k]$ is the peak probability of the normalized context distribution $\bar{\mathbf{P}}_{i,j}^{\,\text{ref}} = \mathbf{P}_{i,j}^{\,\text{ref}} \,/\, \|\mathbf{P}_{i,j}^{\,\text{ref}}\|_1$. The first rule removes contexts whose refined prediction remains normal, suggesting insufficient abnormal evidence under the current model. The second rule discards contexts whose predictions remain ambiguous even after multi-frame aggregation, indicating insufficient evidence to commit to a diagnosis.

\paragraph{Summary assembly.}
For each surviving lesion context, we select the medoid of $\mathcal{H}_{i,j}^{\,\text{ret}}$ in visual feature space as the representative keyframe $\tilde{I}_{i,j}$, providing a visually typical image for physician inspection. The final visual summary contains all surviving contexts after pruning:
\begin{equation}\label{eq:report}
\small
    \mathcal{R}
    = \bigl\{\,(\tilde{I}_{i,j},\;\hat{y}_{i,j})\,\bigr\}\,.
\end{equation}
The number of output summary items is therefore adaptive, matching the variable lesion burden across patients.

\section{Experiments}
\label{sec:experiments}

\subsection{Baselines}

We compare \model~against representative methods under two settings: zero-shot transfer and full training.

In the \emph{zero-shot} setting (Table~\ref{table:zero_shot_main}), we evaluate off-the-shelf models without CE-specific training to measure intrinsic cross-domain generalization. All zero-shot models are combined with ViLAMP frame selection strategy and the same frame budget, so the comparison isolates differences in diagnostic transfer rather than downstream sampling. The compared methods include endoscopy-oriented specialist models (ColonX~\cite{ji_colon-x_2025}, ColonGPT~\cite{colongpt}), whose training data include capsule endoscopy sources such as Kvasir-Capsule, making them among the closest publicly available domain matches to CE, and general medical MLLMs (MedGemma~\cite{medgemma}, HuatuoGPT-Vision~\cite{huatuogptvision}, LingShu~\cite{xu2025lingshu}) trained on broader medical corpora. This setting tests whether either endoscopy-oriented specialist knowledge or generic medical pre-training can transfer to CE.

In the \emph{full-training} setting (Table~\ref{table:main_results}), all methods use the same Selector as a high-recall pre-screening stage, which creates a shared candidate pool before downstream summarization. We then build strong baselines by pairing three backbone families with two frame-selection strategies: general-purpose MLLMs (Qwen3-VL~\cite{qwen3vl}, InternVL3.5~\cite{internvl3.5}), a self-supervised visual backbone (DINOv3~\cite{dinov3}), and the two strongest zero-shot medical models (ColonGPT and LingShu). Each backbone is combined with Adaptive Keyframe Sampling (AKS)~\cite{AKS} and ViLAMP~\cite{vilamp}. All baselines use a fixed budget of 32 selected frames, whereas \model~outputs one representative summary frame per surviving context and therefore produces an adaptive number of selected frames, yielding 29.98 selected frames on average on the test set. 
\begin{figure*}[t]
    \centering
    \subfigure[Keyframe label timeline for a representative patient. The shaded region highlights a short interval with frequent label changes.]{
        \includegraphics[width=0.31\linewidth]{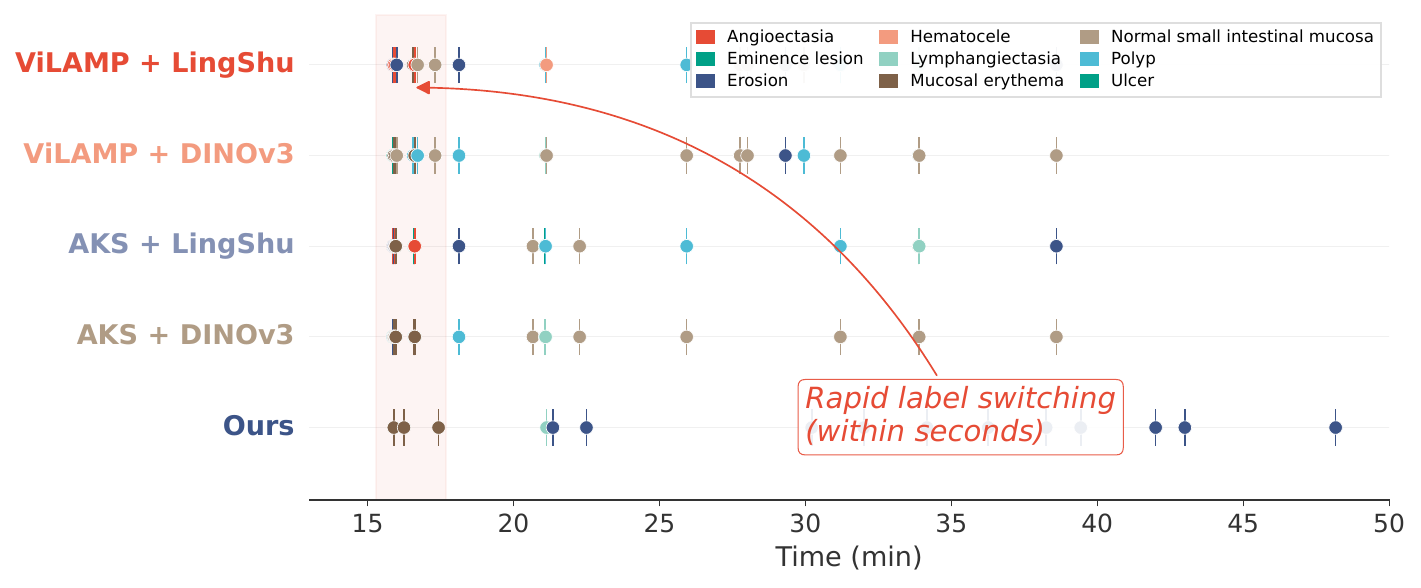}
        \label{fig:uncertain_a}
    }
    \hfill
    \subfigure[Short-range label inconsistency rate at varying temporal thresholds. Lower is better. Error bars denote variation
across patients.]{
        \includegraphics[width=0.31\linewidth]{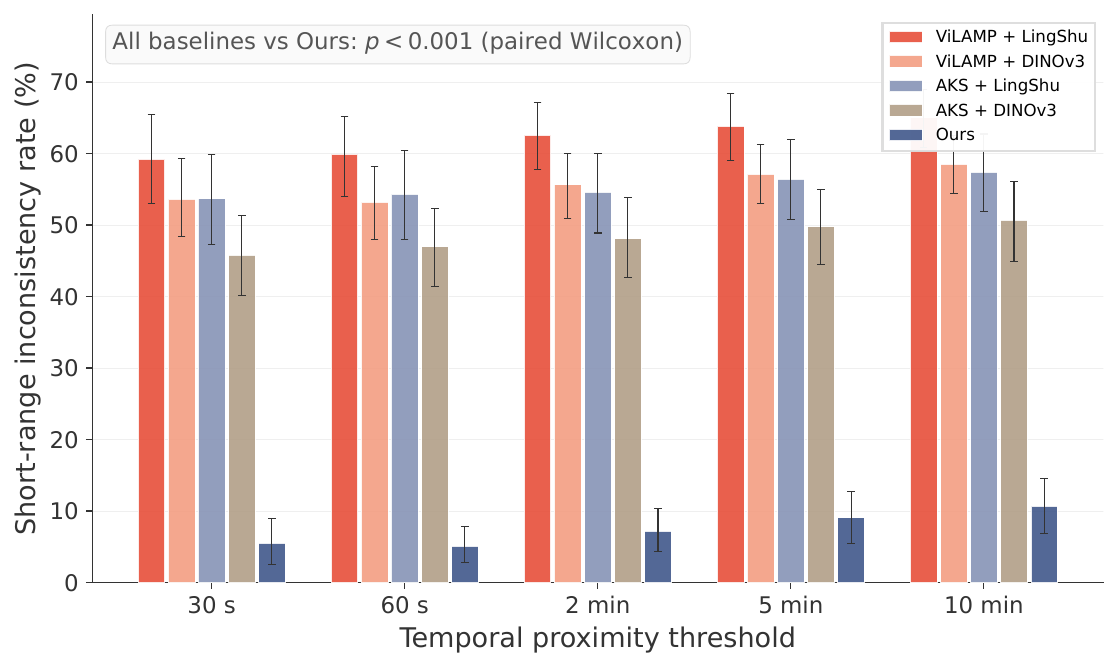}
        \label{fig:uncertain_b}
    }
    \hfill
    \subfigure[Cumulative distribution of label-switch intervals for all
methods.]{
        \includegraphics[width=0.3\linewidth]{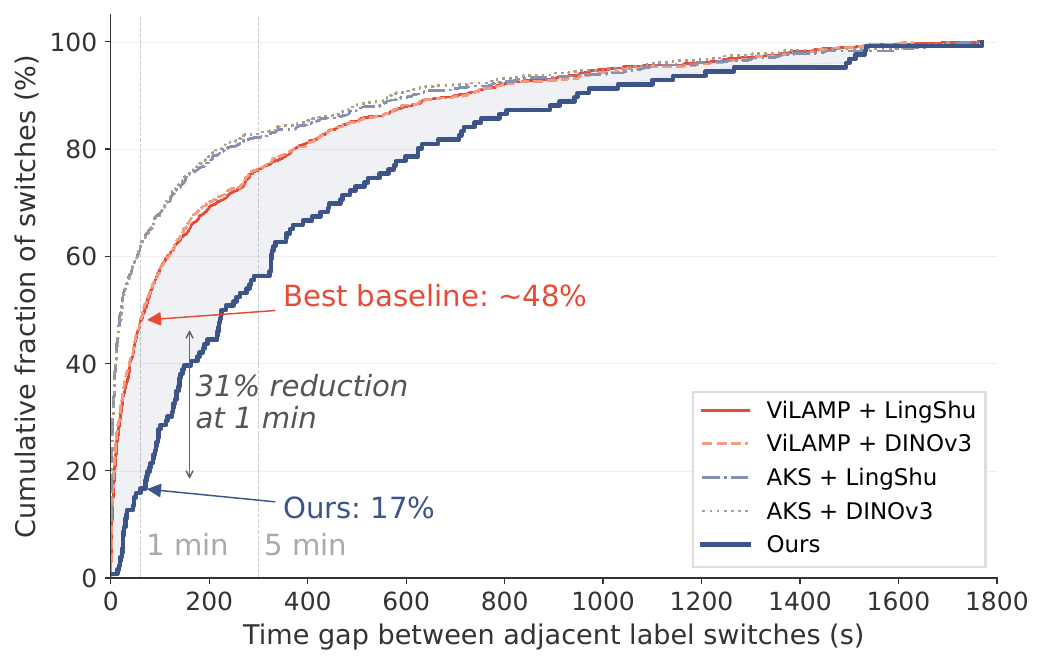}
        \label{fig:uncertain_c}
    }
    \caption{Temporal diagnostic consistency analysis.}
    \label{fig:uncertainty}
\vspace{-5mm}
\end{figure*}

\subsection{Implementation Details}
\label{subsec:implementation}
The Selector uses a frozen DINOv3 encoder with a lightweight MLP head trained by binary cross-entropy on abnormal-versus-normal labels derived from the lesion taxonomy. The diagnoser $g_\phi$ in the Evidence Converger is fine-tuned for 12-class lesion recognition on VideoCAP. At inference time, we fix $\tau_s=0.5$, while $\tau_{\text{agree}}$, $\tau_{\min}$, and the Context Weaver hyperparameters are selected on the validation set.

\subsection{Main Results}
\label{subsec:main_results}

\paragraph{\textbf{Zero-shot models reveal a severe domain gap.}}
The zero-shot results (Table~\ref{table:zero_shot_main}) show a severe domain gap: lesion detection rate ranges from 2.94\% (ColonX) to 17.65\% (LingShu), and no method reaches 20\%. ColonX and ColonGPT achieve the highest sensitivities (72.06\% and 69.85\%) and the lowest time errors (49.91\,s and 54.11\,s), suggesting that endoscopy-oriented pre-training helps temporal localization but remains insufficient for strong CE performance.

\paragraph{\textbf{\model~consistently outperforms state-of-the-art baselines.}}
Table~\ref{table:main_results} shows that \model~achieves the strongest overall performance among fully trained methods. Despite using only 0.2B parameters, it attains the best lesion detection rate (44.12\%), sensitivity (85.29\%), time error, redundancy, diagnostic yield, and patient detection rate. Specificity is 19.30\%, within the baseline range of 18.83--23.55\%. These results suggest that structured context construction and evidence aggregation improve lesion coverage and clinical utility without clearly increasing false positives.

\paragraph{\textbf{Improved keyframe quality and patient-level utility.}}
Beyond lesion coverage, \model~also produces more useful summaries at both the frame and patient levels. It achieves the lowest time error (54.86\,s) and the lowest redundancy (77.81\%), indicating more precise temporal localization and less repeated evidence. At the patient level, \model~achieves the best diagnostic yield (20.00\%) and the best patient detection rate (67.50\%).

  \begin{table}[t]
  \renewcommand{\arraystretch}{1.3}
  \small
  \centering
  \caption{Ablation study on DIV-Cap. For each variant we list the absolute metric value and, on the following line, the absolute difference $\Delta$ relative to the full DIV-Cap
   (\model): $\Delta = v_{\text{variant}} - v_{\text{\model}}$. For percentage metrics $\Delta$ is in percentage points (\%); for Time Error $\Delta$ is in seconds. \textbf{Arrow
   legend:} \textcolor{green!60!black}{$\uparrow$} = performance improved; \textcolor{red}{$\downarrow$} = performance degraded (improvement/deterioration is judged w.r.t.\
  clinical utility; e.g., lower Time Error and lower Redundancy are improvements).}
  \label{table:ablation_divcap}
  \begin{adjustbox}{width=0.8\linewidth}
  \begin{tabular}{l c c c c c c c}
  \toprule
  \multicolumn{1}{c}{} & \multicolumn{3}{c}{\textbf{Lesion-Level}} & \multicolumn{2}{c}{\textbf{Frame-level}} & \multicolumn{2}{c}{\textbf{Patient-Level}} \\
  \cmidrule(lr){2-4} \cmidrule(l){5-6} \cmidrule(l){7-8}
  \multicolumn{1}{c}{\multirow{-2}{*}{\textbf{Variants}}}
  & \textbf{Detection Rate} & \textbf{Sensitivity} & \textbf{Specificity} & \textbf{Time Error} & \textbf{Redundancy} & \textbf{Diagnostic Yield} & \textbf{Detection Rate} \\
   & \textbf{(\%)} & \textbf{(\%)} & \textbf{(\%)}
   & \textbf{(sec)} & \textbf{(\%)}
   & \textbf{(\%)} & \textbf{(\%)} \\
  \midrule
  \midrule
  \textbf{\model}
  & 44.12 & 85.29 & 19.30 & 54.86 & 77.81 & 20.00 & 67.50  \\
  \addlinespace[2pt]
  \midrule
  - Context Weaver
  & 21.36 & 38.83 & 20.00 & 55.09 & 77.18 & 15.00 & 35.00 \\
  \quad $\Delta$ vs. \model
  & \textcolor{red}{$\downarrow$} \(-22.76\ \%\)
  & \textcolor{red}{$\downarrow$} \(-46.46\ \%\)
  & \textcolor{green!60!black}{$\uparrow$} \(+0.70\ \%\)
  & \textcolor{red}{$\downarrow$} \(+0.23\ \text{s}\)
  & \textcolor{green!60!black}{$\uparrow$} \(-0.63\ \%\)
  & \textcolor{red}{$\downarrow$} \(-5.00\ \%\)
  & \textcolor{red}{$\downarrow$} \(-32.50\ \%\) \\
  \addlinespace[4pt]
  - Evidence Converger
  & 18.38 & 29.41 & 15.39 & 75.48 & 84.60 & 2.50 & 27.50 \\
  \quad $\Delta$ vs. \model
  & \textcolor{red}{$\downarrow$} \(-25.74\ \%\)
  & \textcolor{red}{$\downarrow$} \(-55.88\ \%\)
  & \textcolor{red}{$\downarrow$} \(-3.91\ \%\)
  & \textcolor{red}{$\downarrow$} \(+20.62\ \text{s}\)
  & \textcolor{red}{$\downarrow$} \(+6.79\ \%\)
  & \textcolor{red}{$\downarrow$} \(-17.50\ \%\)
  & \textcolor{red}{$\downarrow$} \(-40.00\ \%\) \\
  \bottomrule
  \end{tabular}
  \end{adjustbox}
  \end{table}
\begin{figure*}[t]
\centering
   
    \includegraphics[width=\textwidth]{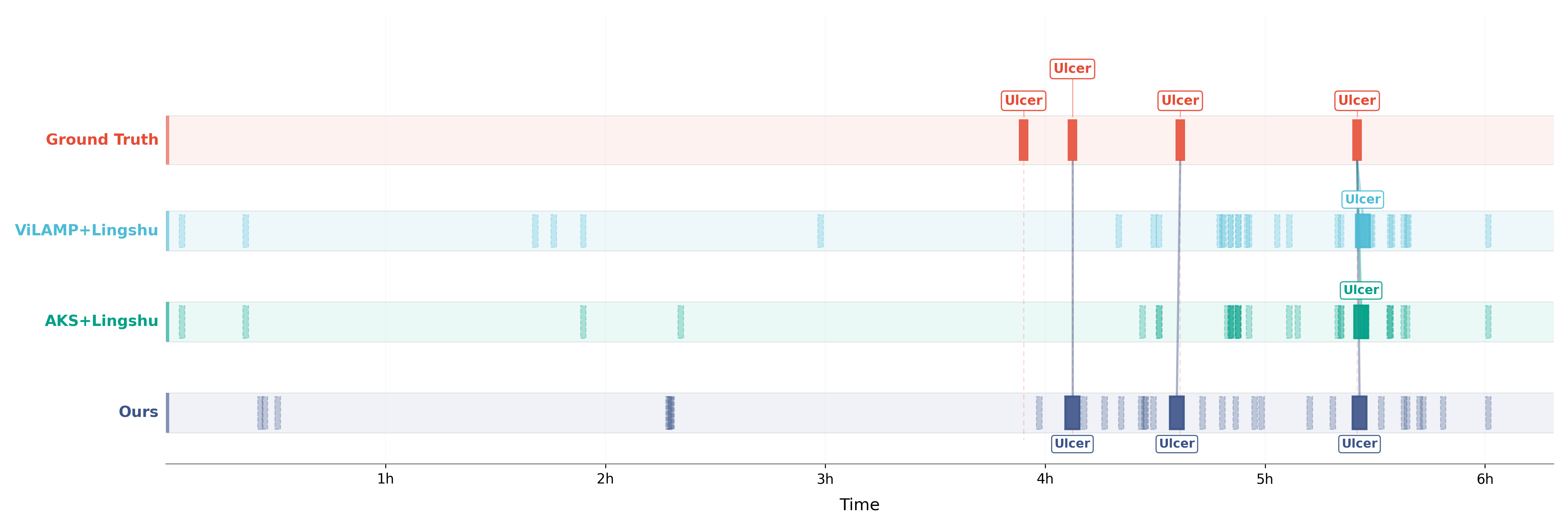}
    \caption{Temporal case study on a representative CE examination.}

\label{fig:case_study}
\end{figure*}
\subsection{Temporal Diagnostic Uncertainty Analysis}
\label{sec:uncertainty}

Temporal diagnostic uncertainty in CE often appears as inconsistent labels across nearby keyframes. We evaluate this phenomenon on the 40-patient test set by comparing \model~with four strong baselines: ViLAMP+LingShu, ViLAMP+DINOv3, AKS+LingShu, and AKS+DINOv3.

\paragraph{\textbf{Metrics.}}
We define the \emph{short-range label inconsistency rate} at threshold $\tau$ as the fraction of consecutive keyframe pairs within each patient whose timestamps differ by at most $\tau$ seconds but receive different labels. Lower values indicate better temporal consistency. We report this metric at $\tau \in \{30\text{s},\; 60\text{s},\; 2\text{min},\; 5\text{min},\; 10\text{min}\}$ and also analyze the cumulative distribution of label-switch intervals.

\paragraph{\textbf{Qualitative evidence (Fig.~\ref{fig:uncertain_a}).}}
Figure~\ref{fig:uncertain_a} shows a representative interval in which baselines switch repeatedly among nearby labels, whereas \model~remains more stable.

\paragraph{\textbf{Short-range inconsistency (Fig.~\ref{fig:uncertain_b}).}}
Across all thresholds, the baselines remain high at 46\%--65\%, meaning that roughly half of nearby keyframe pairs receive conflicting labels. In contrast, \model~reduces the inconsistency rate to 5.1\%--10.7\%, yielding an \textbf{8--9$\times$ reduction} at the 30\,s and 60\,s thresholds (paired Wilcoxon signed-rank test, $p < 0.001$ for all pairwise comparisons at every threshold). The smaller increase with larger $\tau$ suggests that the remaining label changes in \model~are more compatible with genuine temporal transitions than pervasive label noise.

\paragraph{\textbf{Distribution of label-switch intervals (Fig.~\ref{fig:uncertain_c}).}}
Figure~\ref{fig:uncertain_c} further shows that \model~specifically suppresses short-interval switches. For the strongest baseline, ViLAMP+LingShu, about 48\% of label switches occur within 1\,min, versus 17\% for \model. The total number of switch events is also reduced by more than $4\times$ (155 versus 654--832). These results indicate that \model~shifts label changes toward longer intervals and reduces rapid diagnostic oscillation.

\paragraph{\textbf{Clinical implications.}}
Temporally inconsistent summaries increase cognitive burden because clinicians must reconcile conflicting predictions before diagnosis. By producing more coherent label sequences, \model~makes summaries more reliable and easier to interpret.

\subsection{Ablation Study}

Table~\ref{table:ablation_divcap} isolates the contribution of each major module. The full model provides the best balance, while each ablation exposes a distinct failure mode.

\paragraph{\textbf{Without Context Weaver.}}
Replacing the hierarchical Context Weaver with a coarse 300\,s temporal window causes large recall losses: LDR drops by 22.76\%, sensitivity drops by 46.46\%, and patient detection rate drops by 32.50\%. Specificity changes only slightly (+0.70\%), while time error and redundancy remain nearly unchanged. This suggests that temporal proximity alone is insufficient for CE evidence grouping.

\paragraph{\textbf{Without Evidence Converger.}}
Removing the Evidence Converger produces the most severe degradation. When each clip is labeled by its single highest-confidence frame instead of multi-frame aggregation, sensitivity drops by 55.88\%, diagnostic yield drops from 20.00\% to 2.50\%, time error worsens by 20.62\,s, redundancy increases by 6.79\%, and patient detection rate drops by 40.00\%. This shows that single-frame confidence is too unstable for CE diagnosis.

\begin{table}[htbp]
\renewcommand{\arraystretch}{1.3}
\centering
\caption{Selector backbone comparison for binary frame screening.}
\label{table:Selector}
\begin{adjustbox}{width=0.4\linewidth}
\begin{tabular}{lcccc}
\toprule
\textbf{Backbone} & \textbf{Accuracy} & \textbf{Precision} & \textbf{Recall} & \textbf{F1} \\ \midrule
SigLIP2 & 0.8887 & 0.8912 & 0.8887 & 0.8896 \\
DINOv2  & 0.8998 & 0.9069 & 0.8998 & 0.9017 \\
DINOv3  & \textbf{0.9144} & \textbf{0.9145} & \textbf{0.9144} & \textbf{0.9144} \\ \bottomrule
\end{tabular}
\end{adjustbox}
\end{table}

\paragraph{\textbf{Selector backbone choice.}}
Table~\ref{table:Selector} compares three backbones for the Selector. DINOv3 performs best on all four screening metrics, so we adopt it as the Selector backbone.

\subsection{Case Study}

Figure~\ref{fig:case_study} visualizes a representative CE examination with multiple annotated ulcer events. \model~detects more ulcer events and localizes them more accurately in time: it produces three ulcer summaries that align closely with the later three ground-truth timestamps, whereas both ViLAMP+LingShu and AKS+LingShu yield only one clear ulcer summary around the final event and miss the earlier ulcers. This qualitative pattern is consistent with the quantitative results.

\section{Conclusion}

We introduce a new CE video analysis task, diagnosis-driven video summarization, which extracts key evidence frames and diagnoses from them. We propose \model, a novel clinician-inspired framework that mirrors the clinician reading workflow by shifting from frame-level analysis to contextual reasoning. Through its Selector, Context Weaver, and Evidence Converger, \model~first preserves candidate lesion events, then organizes them into diagnostic contexts, and finally aggregates multi-frame evidence into robust clip-level judgments. We further introduce VideoCAP, the first dataset of 240 full-length CE videos with diagnosis-driven annotations derived from real clinical reports, enabling model development and realistic evaluation of CE video summarization. Experiments demonstrate that \model~consistently outperforms baselines,  producing concise and clinically reliable diagnostic summaries. Our results highlight the importance of context-aware evidence aggregation for long CE video understanding and suggest a promising direction toward AI-assisted CE review systems that can reduce clinician workload while preserving diagnostic reliability.

\bibliographystyle{assets/plainnat}
\bibliography{main}

\end{document}